\documentclass[letterpaper, 10 pt, journal, twoside]{ieeetran}  

\usepackage{etex}

\usepackage{multicol}
\usepackage[colorlinks=true, allcolors=blue]{hyperref}

\usepackage{amsmath,amsfonts,amssymb,amsthm} 
\usepackage{float}
\usepackage{bm}
\usepackage{subcaption}
\usepackage{graphicx}
\usepackage{epstopdf}
\usepackage{epsfig}
\usepackage{xspace}
\usepackage{multirow}
\usepackage{hhline}
\usepackage[export]{adjustbox}
\usepackage{csvsimple}

\usepackage[colorinlistoftodos]{todonotes}
\newcommand{\journalVersion}[1]{}
\usepackage{xr}
\usepackage{titlesec}
\graphicspath{{./figures/}}

\usepackage{tabularx}
\newcolumntype{Y}{>{\centering\arraybackslash}X} 

\renewcommand{\baselinestretch}{0.94}

\usepackage{comment}
\usepackage[binary-units]{siunitx}
\usepackage{relsize}
\usepackage{ifthen}
\usepackage[colorinlistoftodos]{todonotes}

\usepackage[vlined,ruled,linesnumbered]{algorithm2e}
\usepackage{graphics} 
\usepackage{rotating}
\usepackage{color}
\usepackage{enumerate}
\usepackage[T1]{fontenc}
\usepackage{psfrag}
\usepackage{epsfig} 
\usepackage{booktabs}
\usepackage{graphicx,url}
\usepackage{multirow}
\usepackage{array}
\usepackage{latexsym}
\usepackage{amsfonts}
\usepackage{amsmath}
\usepackage{amssymb}
\usepackage{xstring}
\usepackage[noend]{algorithmic}
\usepackage{multirow}
\usepackage{xcolor}
\usepackage{prettyref}
\usepackage{flexisym}
\usepackage{bigdelim}
\usepackage{breqn} 
\usepackage{listings}
\let\labelindent\relax
\usepackage{enumitem}
\usepackage{xspace}
\usepackage{bm}
\graphicspath{{./figures/}}
\usepackage{tikz}
\usetikzlibrary{matrix,calc}

\usepackage{mdwlist}

\let\itemize\undefined
\makecompactlist{itemize}{stditemize}

\newcommand{\bdmath}{\begin{dmath}}
\newcommand{\edmath}{\end{dmath}}
\newcommand{\beq}{\begin{equation}}
\newcommand{\eeq}{\end{equation}}
\newcommand{\bdm}{\begin{displaymath}}
\newcommand{\edm}{\end{displaymath}}
\newcommand{\bea}{\begin{eqnarray}}
\newcommand{\eea}{\end{eqnarray}}
\newcommand{\beal}{\beq \begin{array}{ll}}
\newcommand{\eeal}{\end{array} \eeq}
\newcommand{\beas}{\begin{eqnarray*}}
\newcommand{\eeas}{\end{eqnarray*}}
\newcommand{\ba}{\begin{array}}
\newcommand{\ea}{\end{array}}
\newcommand{\bit}{\begin{itemize}}
\newcommand{\eit}{\end{itemize}}
\newcommand{\ben}{\begin{enumerate}}
\newcommand{\een}{\end{enumerate}}

\newcommand{\setal}{~\emph{et~al.}\xspace}
\newcommand{\eg}{\emph{e.g.,}\xspace}
\newcommand{\ie}{\emph{i.e.,}\xspace}
\newcommand{\myParagraph}[1]{{\bf #1.}\xspace}

\newcommand{\M}[1]{{\bm #1}} 
\renewcommand{\boldsymbol}[1]{{\bm #1}}

\newcommand{\hide}[1]{}

\newcommand{\hiddenText}{{\color{gray} hidden text.}}
\newcommand{\hideWithText}[1]{\hiddenText}

\newcommand{\normsq}[2]{\left\|#1\right\|^2_{#2}}

\newcommand{\tran}{^{\mathsf{T}}}

\newcommand{\inv}{^{-1}}

\newcommand{\zero}{{\mathbf 0}}
\newcommand{\eye}{{\mathbf I}}

\newcommand{\matTwo}[1]{\left[\begin{array}{cc}  #1  \end{array}\right]}

\newcommand{\SEthree}{\ensuremath{\mathrm{SE}(3)}\xspace}

\newcommand{\MM}{\M{M}}

\newcommand{\MT}{\M{T}}

\newcommand{\MOmega}{\M{\Omega}}

\newcommand{\vv}{\boldsymbol{v}}
\newcommand{\vt}{\boldsymbol{t}}

\newcommand{\blue}[1]{{\color{blue}#1}}

\newcommand{\linkToPdf}[1]{\href{#1}{\blue{(pdf)}}}
\newcommand{\linkToPpt}[1]{\href{#1}{\blue{(ppt)}}}
\newcommand{\linkToCode}[1]{\href{#1}{\blue{(code)}}}
\newcommand{\linkToWeb}[1]{\href{#1}{\blue{(web)}}}
\newcommand{\linkToVideo}[1]{\href{#1}{\blue{(video)}}}
\newcommand{\linkToMedia}[1]{\href{#1}{\blue{(media)}}}
\newcommand{\award}[1]{\xspace} 

\newcommand{\vislam}{VI-SLAM\xspace}

\newcommand{\bmat}{\left[ \begin{array}}
\newcommand{\emat}{\end{array} \right]}

\newcommand{\bal}{\begin{align}}
\newcommand{\eal}{\end{align}}

\titlespacing*{\section}{0pt}{4mm}{2mm}
\titlespacing*{\subsection}{0pt}{2mm}{2mm}

\newcommand{\specialcell}[2][c]{%
  \begin{tabular}[#1]{@{}c@{}}#2\end{tabular}}

\usepackage{siunitx}
\newcommand{\isExtended}[2]{#2} 

\title{
  \huge{Multi-Camera Visual-Inertial Simultaneous Localization \\ and Mapping for Autonomous Valet Parking}
}

\author{Marcus Abate$^1$, 
Ariel Schwartz$^2$,
Xue Iuan Wong$^3$,
Wangdong Luo$^3$,
\\
Rotem Littman$^2$,
Marc Klinger$^3$,
Lars Kuhnert$^3$,
Douglas Blue$^3$, 
Luca Carlone$^1$
\vspace{-3mm}
\thanks{
$^1$The authors are with the Laboratory for 
Information \& Decision Systems (LIDS), Massachusetts Institute of Technology, Cambridge, USA, 
{\tt\scriptsize \{mabate,lcarlone\}@mit.edu}
}
\thanks{
$^2$The authors are with SAIPS, \mbox{Tel Aviv, Israel, {\tt\scriptsize \{ariel\_schwartz,}}
 {\tt\scriptsize rotem\_littman\}@saips.co.il}
}
\thanks{
$^3$The authors are with Ford Motor Company, Dearborn, Michigan, USA, 
{\tt\scriptsize \{wluo1,xwong,lkuhnert,mklinger,dblue\}@ford.com}
}
\thanks{This work was partially funded by Ford Motor Company.
}
}

\begin{document}


\maketitle

\begin{tikzpicture}[overlay, remember picture]
\path (current page.north east) ++(-6.3,-0.0) node[below left] {
Accepted for publication at ISER 2023, please cite as follows:
};
\end{tikzpicture}
\begin{tikzpicture}[overlay, remember picture]
\path (current page.north east) ++(-3.5,-0.4) node[below left] {
  M. Abate, A Schwartz, X Iuan Wong, W Luo, R Littman, M Klinger, L Kuhnert, D Blue, L Carlone
};
\end{tikzpicture}
\begin{tikzpicture}[overlay, remember picture]
\path (current page.north east) ++(-3.5,-0.8) node[below left] {
  ``Multi-Camera Visual-Inertial Simultaneous Localization and Mapping for Autonomous Valet Parking'',
};
\end{tikzpicture}
\begin{tikzpicture}[overlay, remember picture]
\path (current page.north east) ++(-7.5,-1.2) node[below left] {
  IEEE Int. Conf. Exp. Robotics. (ISER), 2023.
};
\end{tikzpicture}

\vspace{-5mm}
\begin{abstract}
    
    Localization and mapping are key capabilities for self-driving vehicles. 
    In this paper, 
    we build on Kimera and extend it to use multiple cameras as well as external (\eg wheel)  odometry sensors, to obtain accurate and robust odometry estimates in real-world problems.
    Additionally, we propose an effective scheme for closing loops that circumvents the drawbacks of common alternatives based on the Perspective-n-Point method and also works with a single monocular camera.
    Finally, we develop a method for dense 3D mapping of the free space that combines a segmentation network for free-space detection with a homography-based dense mapping technique.
    We test our system on photo-realistic simulations and on several real datasets collected on a car prototype developed by the Ford Motor Company, spanning both indoor and outdoor parking scenarios.
    Our multi-camera system is shown to outperform state-of-the art open-source visual-inertial-SLAM pipelines (Vins-Fusion, ORB-SLAM3), and exhibits an average trajectory error under 1\% of the trajectory length 
    across more than \SI{8}{\km} of distance traveled (combined across all datasets).
    A video showcasing the system is available at: \href{https://youtu.be/H8CpzDpXOI8}{youtu.be/H8CpzDpXOI8}
    
\end{abstract}
    


\section{Introduction}
\label{sec:introduction}

Visual-inertial (VI) SLAM algorithms have seen widespread use in a variety of robotics platforms, from drones to rockets and ground robots~\cite{Forster17tro,Mourikis09tro-EdlSoundingRocket,Cadena16tro-SLAMsurvey}.  
Typically, these algorithms employ a monocular or stereo camera and an inertial measurement unit  
(IMU). Using a single camera works well for aerial vehicles and small robots, where payload constraints limit the number of onboard sensors. However, in other applications, such as self-driving cars, one would prefer to 
use multiple cameras around the vehicle to improve accuracy and robustness of visual-inertial SLAM and enable a broader coverage for 3D mapping of the vehicle's surroundings. 

In this paper we study multi-camera \vislam for autonomous valet parking.
Vision-based autonomous parking is generally less well-studied than highway or city driving in the research literature, and  presents unique challenges. 
Parking can happen in outdoor environments with many dynamic obstacles such as pedestrians or other cars, which necessitates very accurate free-space mapping for safe navigation. 
Parking can also happen in indoor parking garages, which are generally GPS denied environments with visually similar scenes throughout (\eg think about the different floors of an indoor parking garage), making place recognition and drift correction difficult. 
Additionally, parking scenarios see the car traveling at low speeds for long stretches, often creating degenerate conditions for visual-inertial odometry estimation.
A comprehensive autonomous parking solution must be accurate and robust both in state estimation and mapping, and must function in GPS-denied environments, while simultaneously being capable of accepting other sensor inputs when available. 
%
There exist several efforts in the literature for performing multi-camera SLAM with IMUs. 
Of the methods that have been evaluated on datasets targeted at the autonomous parking problem, most of the approaches build 2D representations of the environment, which limits their applicability to multi-story parking garages~\cite{Shao20acm-ATS, Shao22acm-SLAMFI, Yu22ieee-HierarchicalMI, Xiang21icra-HybridBirdsEyeValet}. 
Methods that do build 3D maps either require sensor fusion with more expensive sensors (\eg LiDAR)~\cite{Khoche22icits-Semantic3G, Shi22-VIPSRP} or are not close to real-time operation. 

In this system paper, we develop a multi-camera \vislam pipeline that can perform efficient and globally consistent trajectory estimation and builds a dense 3D map of the free space around the vehicle, which 
enables obstacle avoidance and navigation.
The proposed system builds on Kimera~\cite{Rosinol21ijrr-Kimera,Rosinol20icra-Kimera} and extends it to (i) accept
 multi-camera and external odometry sources, (ii) enable robust monocular or multi-camera loop closures,
 and (iii) perform efficient ground-plane mapping for autonomous valet parking applications. 
Kimera's frontend and backend are modified to improve tracking and factor-graph optimization, and support multi-sensor fusion, using a heavily parallelized architecture. 
Several loop-closure methods are implemented including monocular loop closure techniques that are shown to outperform popular approaches based on the Perspective-n-Point (PnP) method.
 Finally, our SLAM system uses Kimera-Semantics~\cite{Rosinol21ijrr-Kimera,Rosinol20icra-Kimera} in conjunction with a fast semantic segmentation network to create a 3D map of the free space around the robot. 
The method is validated in photo-realistic simulations and on several real datasets collected using a car prototype developed by the Ford Motor Company, spanning both indoor and outdoor parking scenarios.
Our multi-camera system is shown to outperform state-of-the art open-source visual-inertial-SLAM pipelines (Vins-Fusion, ORB-SLAM3), and exhibits an average trajectory error under 1\%  of the trajectory length
across more than \SI{8}{\km} of distance traveled (combined across all datasets).

This paper is organized as follows. 
Section \ref{sec:relatedWork} discusses relevant literature. 
Section \ref{sec:systemarchitecture} describes the proposed method. 
Section \ref{sec:experiments} details experimental setup and results, and Section~\ref{sec:conclusions} concludes the paper.

\begin{figure*}[ht]
\vspace{-6mm}
\centering
\begin{subfigure}[b]{0.35\textwidth}
    \centering
    \includegraphics[width=\textwidth]{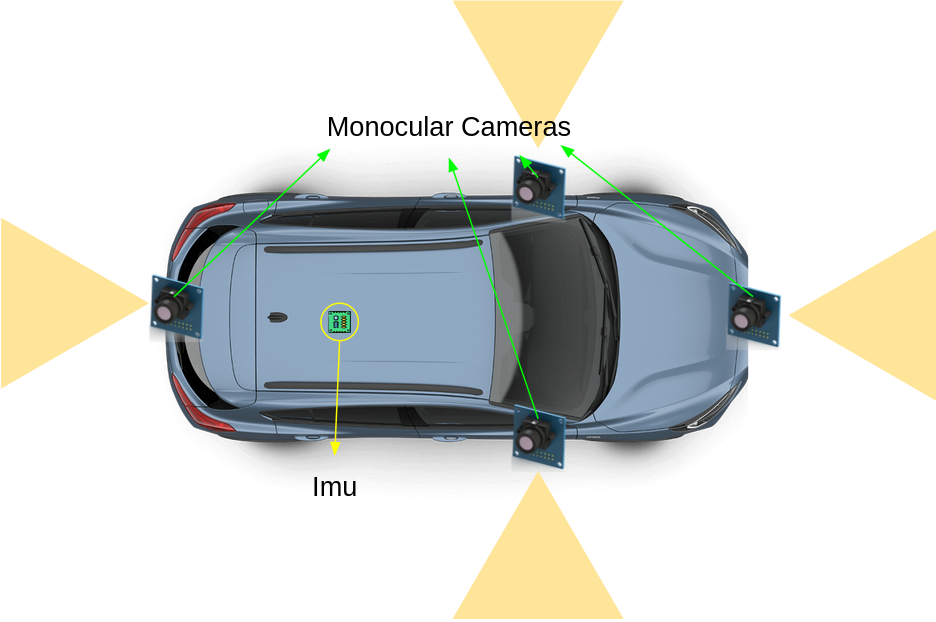}
    \caption{}
    \label{fig:dataset-car}
\end{subfigure}
\hfill
\begin{subfigure}[b]{0.35\textwidth}
    \includegraphics[width=\textwidth]{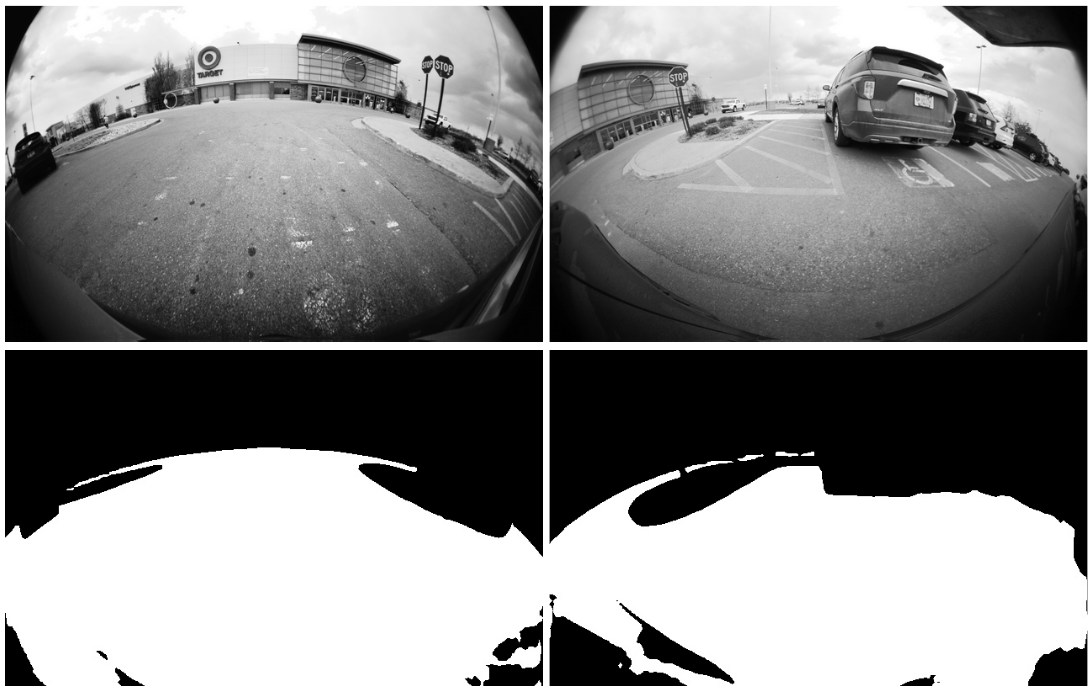}
    \caption{}
    \label{fig:dataset-pics}
\end{subfigure}
\hfill
\begin{subfigure}[b]{0.28\textwidth}
    \includegraphics[width=\textwidth]{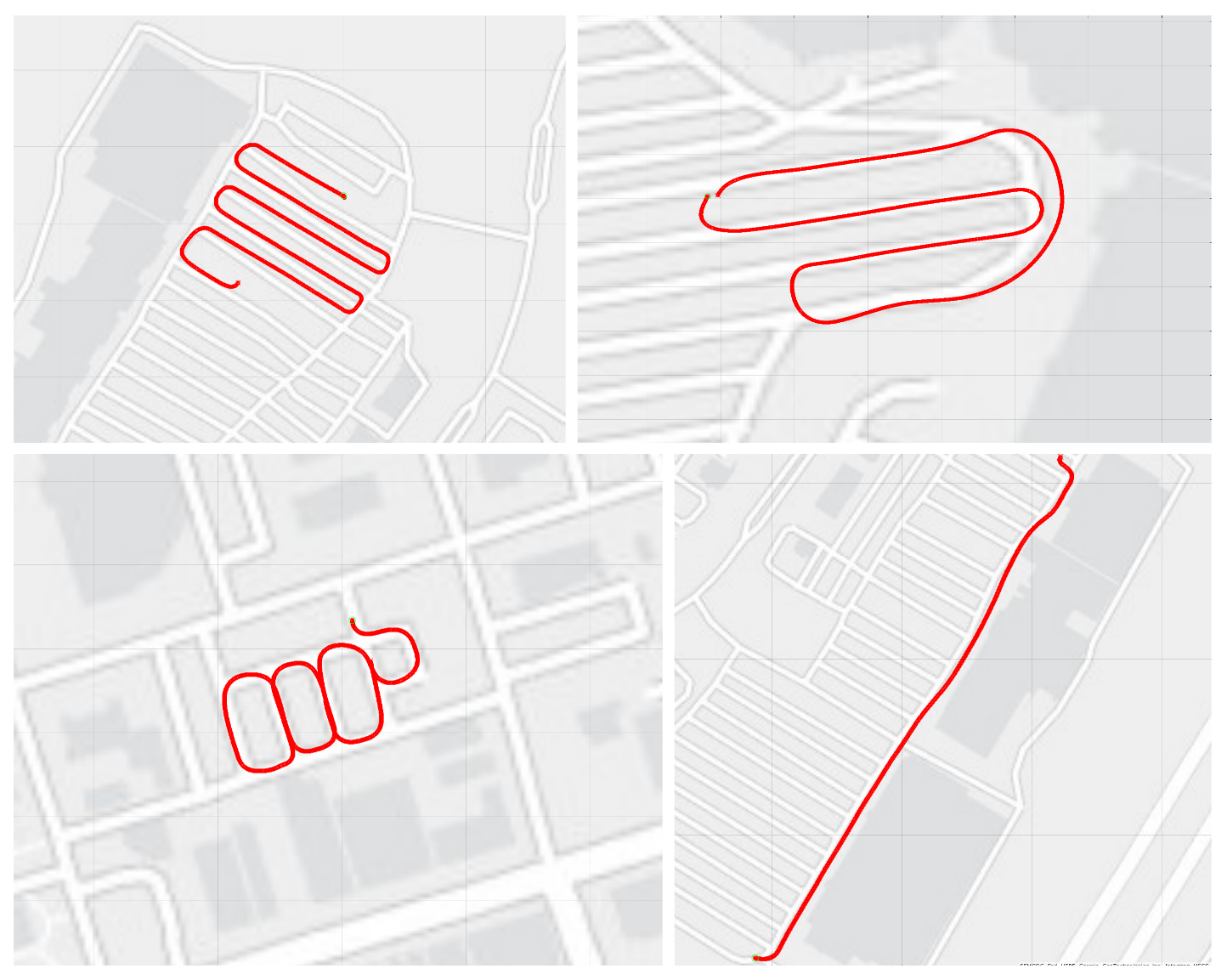}
    \caption{}
    \label{fig:dataset-traj}
\end{subfigure}
\caption{
    (a) Illustration of the Ford test bed and sensor setup.
    (b) Four sample images from an outdoor dataset; the top two are from the front and right cameras onboard the car and the bottom two are the output of the semantic segmentation network that identifies free-space road for the mapping module. 
    (c) Sample of outdoor trajectories collected on the car in Detroit, Michigan, USA.
    The pictured trajectories are on average \SI{450}{\m} in length.
    }
\label{fig:dataset}
\vspace{-5mm}
\end{figure*}

\section{Related Work}
\label{sec:relatedWork}



\myParagraph{\vislam Systems} 
Previous-generation open-source \vislam algorithms, such as Vins-Mono \cite{Qin18tro-vinsmono} and ORB-SLAM2 \cite{Mur-Artal17tro-ORBSLAM2}, used monocular camera and IMU input.
Their modern counterparts support stereo cameras for accurate depth estimation, including Vins-Fusion \cite{Qin19arxiv-VINS-Fusion-poseEstimation}, ORB-SLAM3 \cite{Campos21-TRO}, Open-VINS \cite{Geneva20icra-openVINS}, and Kimera \cite{Rosinol21ijrr-Kimera}. 
At a high-level, the structure of these systems is fairly similar, with a front-end module that performs 
feature detection and tracking (or matching), and a back-end that estimates the trajectory and a sparse landmark-based map via factor graph optimization or an Extended Kalman Filter; 
the front-end is typically based on OpenCV~\cite{OpenCV} for image processing, while the backend is commonly based on optimization libraries, \eg Kimera uses GTSAM \cite{gtsam}, while Vins-Fusion uses Ceres \cite{Agarwal12-ceres}.
Generally speaking, these backend architectures are flexible enough to accept different factors modeling different types of sensor data. 
This has led to novel approaches for sensor fusion that combine visual-inertial or inertial factors with other sensors like LiDAR, such as~\cite{Shi22-VIPSRP} and~\cite{Chang22ral-LAMP2}. 

\myParagraph{Multi-Camera \vislam} 
One option for increasing the estimation accuracy and robustness of \vislam methods is to 
leverage multiple cameras mounted on the robot. 
Frequently in the literature, the use of multiple cameras arranged around the robot is referred to as ``surround-view''. 
Eckenhoff\setal~\cite{Eckenhoff20arxiv-MIMICVINS} implement an EKF-based visual-inertial odometry (VIO) pipeline that supports multiple cameras and IMUs. 
Yang\setal~\cite{Yang21arxiv-AsyncMultiViewSLAM} use several pinhole cameras with fixed intrinsics and extrinsics. 
Zhang\setal~\cite{Zhang21ral-BalancingTB} improve on these concepts by using a single set of feature tracks among all cameras (with overlapping fields-of-view), along with a factor-graph-based backend. 
Zhang\setal~\cite{Zhang22irs_MMOSlam} implement a multiple-monocular SLAM system with online extrinsic calibration, using an ORB-based frontend and wheel-odometry for initialization. 
He\setal~\cite{He21arxiv-TowardEA} also implement cross-camera feature sharing without constraints on the camera field of view to produce notable results, but at the cost of requiring GPU acceleration for the frontend and bundle-adjustment steps. 
Wang\setal~\cite{Wang20icra-ReliableFM} implement an efficient scheme for surround-view localization using 4 cameras on a ground robot by assuming planar motion only. 
The distinguishing features of our system are its capability to fuse multiple asynchronous camera feeds, support robust monocular loop closure optimization, and perform 
free-space mapping using monocular cameras, all without relying on specialized hardware beyond the CPU.

\myParagraph{Autonomous Valet Parking} 
Several recent works have tackled the specific problem of autonomous parking in the context of self-driving vehicles. 
Tripathi and Yogamani~\cite{Tripathi20arxiv-TrainedParkingSurroundView} summarize the challenges of using visual-inertial odometry to build global maps for relocalization, which is a necessary component of real-time autonomous parking. 
Shao\setal~\cite{Shao20acm-ATS,Shao22acm-SLAMFI} introduce a VIO system that leverages surround-view cameras, but for the purpose of detecting parking spots instead of in the SLAM loop. 
Yu\setal~\cite{Yu22ieee-HierarchicalMI} and Xiang\setal~\cite{Xiang21icra-HybridBirdsEyeValet} use surround-view images in a similar way, but take the extra step of performing feature tracking on these bird's-eye view images and incorporating those features in a hierarchical factor-graph optimization.
Performing SLAM on surround-view images in this way comes at the cost of assuming strictly planar movement on a 2D map, which is sufficient for ground vehicles in most cases, but limits performance in multi-story parking lots and limits extensibility to future applications that require 3D navigation (\eg off-road navigation). 
These limitations are partially addressed by Khoche\setal~\cite{Khoche22icits-Semantic3G}, where the authors also propose a 3D mapping method that remains efficient but requires LiDAR. 
Our proposed method does not make planar assumptions but rather tracks movement in 3D, yet makes use of multiple cameras for  improving localization and mapping. 

\myParagraph{Free-Space Mapping for Autonomous Valet Parking} 
Dense maps of the environment are generated by the SLAM algorithm, however at a bare minimum they must be annotated to show free-space in the environment for any practical use (\eg path planning). 
In the case of self-driving cars, free-space is empty road and empty parking spots.
Shao\setal~\cite{Shao20acm-ATS,Shao22acm-SLAMFI} 
use surround-view cameras to generate a planar ground-plane map around the vehicle, and used this map in conjunction with a CNN to detect parking spaces for high-level planning. 
Tripathi and Yogamani~\cite{Tripathi20arxiv-TrainedParkingSurroundView} and MOFISSLAM~\cite{Shao22CSVT-MOFISSLAMAM} both use semantic segmentation networks to identify common labels like pedestrians, speed bumps, etc. 
Wu\setal~\cite{Wu22ieee-VinsDynamicObjDet} use an object detector in the loop with VIO to identify dynamic obstacles and remove features on those obstacles from the visual SLAM frontend. 
%
Building these semantic maps requires a segmentation network to classify free-space (ground-plane), but it also requires a way to get a dense depth map of image pixels that are on the ground. 
A popular solution has been to use mono-depth estimation networks~\cite{Wimbauer20arxiv-MonoRec}. 
An alternative solution in the vein of sensor fusion is to use LiDAR data to gather depth information and align it with the semantically segmented image~\cite{Khoche22icits-Semantic3G}. 



\section{System Architecture}
\label{sec:systemarchitecture}

This section describes the hardware (Section~\ref{subsec:hardware-architecture}) and software architecture  (Section~\ref{subsec:software-architecture}) of the proposed system.

\subsection{Hardware Architecture and Data Collection}
\label{subsec:hardware-architecture}

\myParagraph{Hardware Architecture}
The real-world platform used for experiments was a modified Lincoln MKZ sedan with custom engine control units and custom onboard sensors, including four monocular fisheye cameras, an IMU, and an onboard wheel-odometry system that uses wheel-encoders as well as other proprietary sensors to estimate the car’s motion. 
The sensors are all production-equivalent except for the IMU. 
Figure~\ref{fig:dataset} shows the arrangement of the sensor suite on the car. 
The IMU data was provided by an RT3000 unit, and the cameras were 1-MegaPixel production automotive cameras. 
IMU data was collected at 100Hz while camera data at 20Hz. 
Wheel odometry data was also provided at IMU rate. 
Ground-truth data (only used for benchmarking) was provided by differential GPS for the outdoor datasets at IMU rate. 
All sensors used the same onboard clock but were not synchronized. 
Data was collected via a ROS node developed to interface with the raw sensor data, which were communicated via CAN bus and UDP onboard the car. 
While the car was capable of running Kimera online, the results presented in this paper were obtained using a desktop computer running Ubuntu 20.04 with a 24-core Intel processor. 

\myParagraph{Data Collection}
For evaluation, we used both simulated datasets and real datasets collected with the vehicle described above.
%
The simulated scene was an outdoor urban area, and the data came from a simulated car with a realistic dynamics model. 
Ground-truth pixel-wise semantic labels were extracted from the simulator in place of a segmentation network, and the labels associated with roads were used for the free-space reconstruction in Kimera-Semantics. 

For the real-world datasets, we collected 22 datasets recorded in 5 different locations, both indoor and outdoor, over the course of 5 months. 
The speeds of the car in the datasets varied, as did trajectories, environment, obstacles, and weather (see sample trajectories in Fig.~\ref{fig:dataset-traj}). 
Some of the datasets had occlusions on one or multiple cameras for extended periods, others had long stops for traffic, crowded markets with many pedestrians, and long straight-line motion beyond the VIO module's time horizon that made scale estimation difficult. 
For the indoor datasets, GPS was unreliable inside of the parking garages, so we used Ford's proprietary wheel-odometry  motion estimate as ground truth as it proved extremely accurate, especially at low and medium speeds.  


\begin{figure*}[ht]
\centering
\vspace{-10mm}
\includegraphics[width=0.8\textwidth]{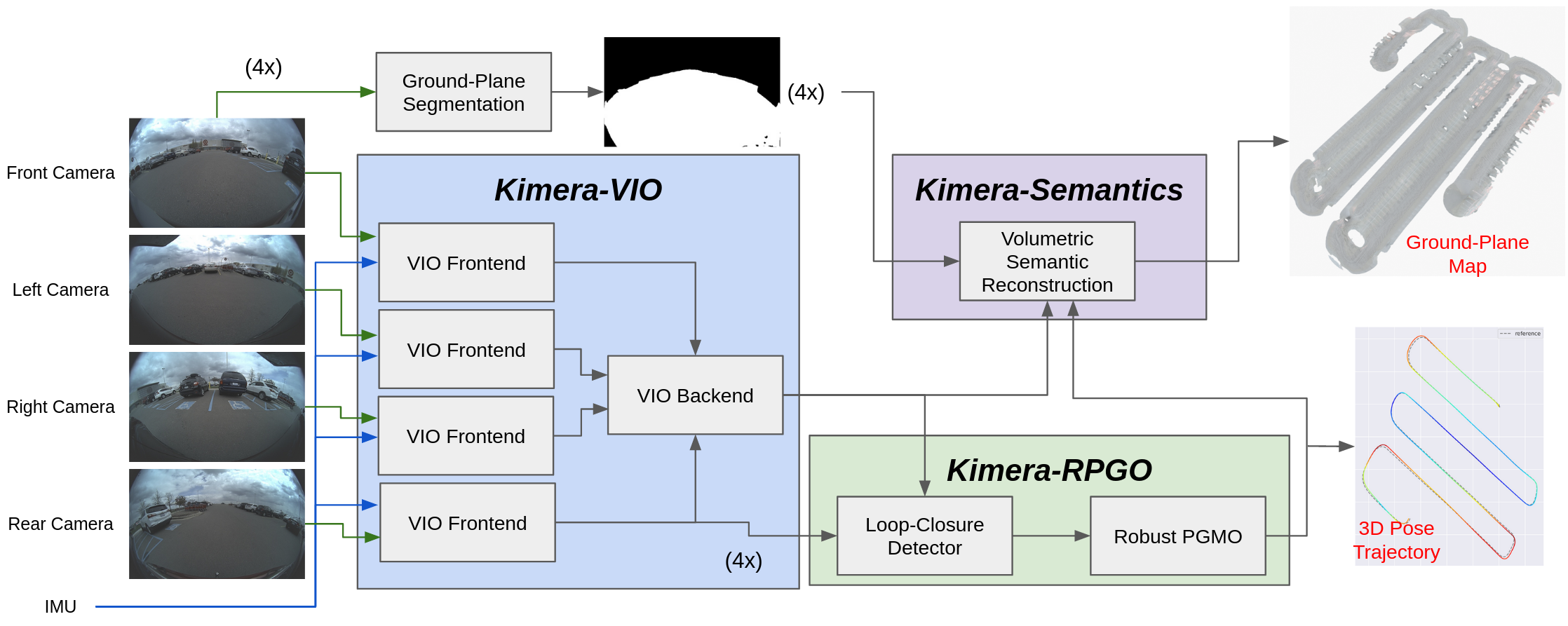}
\caption{Overview of the proposed system architecture. 
Inputs are RGB monocular images from all four sides of the car, as well as a single IMU. 
Our modified Kimera-VIO processes all camera inputs in parallel and generates a robust state estimate, which is fed to the Robust Pose Graph Optimization (RPGO) module for loop closure detection and correction.
Simultaneously, a semantic segmentation network identifies the ground plane in the image, which is used by the modified Kimera-Semantics module to generate a 3D reconstruction of the free space.
 For a more in-depth description of Kimera's modules, refer to \cite{Rosinol21ijrr-Kimera}. 
}
\label{fig:architecture}
\vspace{-5mm}
\end{figure*}

\subsection{Software Architecture}
\label{subsec:software-architecture}


Figure~\ref{fig:architecture} reports the overall architecture of the proposed system, including the \vislam system and the free-space mapping module. 

\myParagraph{Multi-Camera Kimera-VIO}
Previously, Kimera was limited to perform tightly-coupled visual-inertial odometry using a stereo or RGB-D camera.
The proposed version of Kimera is modified to accept monocular image data ---potentially from multiple cameras--- coupled with IMU data. 
The Monocular-VIO frontend is split into two; IMU data is preintegrated with interpolation on both endpoints according to standard methods~\cite{Forster15rss-imuPreintegration}.
The image data is then processed using tools from OpenCV \cite{OpenCV} to detect features (goodFeaturesToTrack) and track them across frames. 
5-point RANSAC is then applied on the tracked features to remove outliers at each keyframe; keyframes are triggered depending on the number and quality of the tracked image features.
In addition to visual and inertial data, we use also wheel odometry measurements.
The relative pose between two frames from the wheel odometry data (multiple measurements between each keyframe-pair are chained together) is used with a tunable noise model as a factor in the VIO backend's factor-graph.
The features, preintegrated IMU measurements, and odometry measurements are then sent to the VIO backend module, which performs fixed-lag smoothing using all the available measurements.


In order to take full advantage of the surround-view camera setup onboard the self-driving car platform, Kimera was  modified to accept any number of monocular or stereo camera inputs.
In the multi-camera configuration, each camera is processed using its own frontend module, and all frontends are run in parallel.
The frontend modules then send their outputs to a single backend, which in turn keeps track of which camera provides which factors and performs a single factor-graph optimization. 
Since the cameras have a wide field of view and have a large distortion (beyond what can be captured by standard distortion models in OpenCV), the factors used in the factor-graph optimization for visual features were modified to increase robustness.
In the standard version of Kimera, GTSAM's {\tt SmartStereoProjectionFactor} is used for each of the visual landmarks \cite{gtsam}.
For the proposed method, the triangulation occurring in the {\tt SmartStereoProjectionFactor} 
  was modified to optionally use the Huber norm to increase robustness to outliers. 
%
As part of this effort, we have also implemented several novel factors in GTSAM, including factors for automatic extrinsic camera calibration, 
rolling shutter correction, 
and projection factors that use the spherical camera model proposed by Scaramuzza\setal~\cite{Scaramuzza06iros-omniCamera}. We  have omitted them from this paper since we haven't seen them produce more accurate results in our tests. 

\myParagraph{Loop Closure Optimization}
While the VIO backend produces a locally consistent trajectory, our goal is to obtain a \emph{globally} consistent estimate of the trajectory and the map. The proposed architecture passes the VIO motion estimates to a robust pose graph optimization module that detects loop closures and optimizes the trajectory accordingly. 
For loop-closure detection, Kimera uses visual Bag-of-Words to detect similar images.
This is done within the scheme described in \cite{Rosinol21ijrr-Kimera} and \cite{Galvez12tro-dbow}.
Once a pair of putative matching images is identified, Kimera must generate a relative pose between the two frames before the factor can be included in the pose-graph optimization.
Below we describe three approaches to compute the loop closure pose.
After the relative pose between frames is computed, it is passed to Kimera-RPGO, for pose graph optimization. In Kimera-RPGO, we use graduated non-convexity~\cite{Yang20ral-GNC} to robustify the estimate to spurious loop closures.

\emph{PnP Loop Closure Pose Computation.}
In the PnP approach, ORB descriptors are extracted at each keyframe and associated to each tracked feature point. 
At the same time, 3D landmark data from the VIO backend is sent to the loop-closure-detection module alongside each image.
We use the ORB descriptors to obtain putative correspondences between the optimized landmarks from the backend and the 2D features, and 
use the PnP algorithm with RANSAC to find inlier correspondences.
While this approach is fairly popular and used in other pipeline (\eg~\cite{Campos21-TRO}), our experiments show that the resulting poses are not very accurate, 
likely due to the sparsity of the 3D landmark-based map, and hence a lack of inliers.

\emph{Scale-less Loop Closure Pose Computation.}
In order to alleviate the problems with the PnP approach, we instead compute a pose up to scale. 
The pose is obtained using a standard 5-point RANSAC method, which can now directly rely on the many 2D-2D correspondences established between image features. 
As we do not have the scale factor on the translation in this case, we modify the information matrix of the noise model associated with the loop closure factor to use only the direction and not the magnitude of the translation vector.
The pose-graph optimization then uses only the rotation part and the translation's direction of the relative pose in the loop closure factor. 
More formally, the scale-less relative pose factors take the following form: 
\beq
\label{eq:betweenFactor}
f(\MT_i, \MT_j) = \normsq{ \text{Log}( \bar{\MT}_{ij}\inv \MT_i\inv \MT_j ) }{\MOmega_{ij}}
\eeq
where the loop closure measurement $\bar{\MT}_{ij} \in \SEthree$ relates two poses $\MT_i, \MT_j \in \SEthree$ along the trajectory of the car, and $\text{Log}(\cdot)$ is the standard logarithm map. 
We then set the information matrix $\MOmega_{ij}$ to be:
\beq
\label{eq:info}
\MOmega_{ij} = 
\matTwo{ \MOmega_R & \zero_{3\times 3} \\ \zero_{3\times 3} & \eye_3 - \bar{\vt}_{ij} \bar{\vt}_{ij}\tran  }
\eeq
where $\MOmega_R$ is the $3\times 3$ information matrix describing the uncertainty in the relative rotation component of the loop closure pose $\bar{\MT}_{ij}$, and $\bar{\vt}_{ij}$ is the translation direction (assumed to be a unit-norm vector) in $\bar{\MT}_{ij}$. The matrix $\MM = \eye_3 - \bar{\vt}_{ij} \bar{\vt}_{ij}\tran$ is the orthogonal projector of the vector $\bar{\vt}_{ij}$ and is such that, for any 3D vector $\vv$, $\MM \vv = 0$ if $\vv$ is aligned with $\bar{\vt}_{ij}$. 
Intuitively, the matrix simply disregards the component of the translation error in the direction of $\bar{\vt}_{ij}$. 

\emph{Rotation-only Loop Closure Computation.}
This last variant is similar to the scale-less approach, but it disregards the translation component and only uses the relative rotation (computed via the 5-point method) of the loop closure.
In our implementation, we still use the factor~\eqref{eq:betweenFactor} but set the translation information matrix 
---\ie the bottom-right block in~\eqref{eq:info}--- to zero. 

\myParagraph{Free-space Mapping via Kimera-Semantics}
In our previous work~\cite{Rosinol21ijrr-Kimera,Rosinol20icra-Kimera}, we used Kimera-Semantics to generate semantically annotated 3D meshes from stereo/depth data. 
In this case, we do not have enough overlapping field of view between cameras to perform stereo reconstruction. 
As we are only concerned with free-space mapping and since the road can be assumed to be locally planar,
we first detect the ground plane in the image using a CNN for pixel-wise binary classification, and then map the ground plane to a 3D plane using a homography transformation~\cite[Chapter 13]{Hartley00}.
 Using the homography, we map every pixel belonging to the ground plane to a 3D point and then pass the corresponding 3D point cloud to Kimera-Semantics, which performs ray-casting to infer a 3D voxel-based map, and then extracts a textured 3D mesh via marching cubes.


\section{Experiments}
\label{sec:experiments}





This section showcases the effectiveness of each component of the proposed system, including the 
visual-inertial odometry (Section~\ref{subsec:pose-estimation}), the loop closure modeling (Section~\ref{subsec:lc-detection}), and the free-space mapping (Section~\ref{subsec:free-space-recon}). 
The results are also compared against state-of-the-art VI-SLAM methods. 
%

\subsection{Visual-Inertial Odometry}
\label{subsec:pose-estimation}

Table \ref{tab:kimera_mono_ate_real} shows the performance of our monocular extension of Kimera-VIO, comparing the performance of each camera in isolation.
We ran each dataset using only one camera at a time. 
Both trajectory ATE RMSE and drift are reported for each configuration and each dataset, and the best results for each dataset are highlighted in green. 
The table covers both simulated datasets (``sim'') as well as real outdoors (``out'') and indoor (``in'') datasets.
Overall, the rear camera performed poorly as compared to the rest of the cameras, likely due to the fact that the camera was tilted up slightly and therefore the (untrackable) sky/ceiling took up most of the image. 
The front camera had the highest number of best-scores on the real datasets, while the left and right cameras were superior in simulation and performed reasonably well on the real datasets. 
The left and right cameras were mounted on the car's side mirrors, which were not stable mounting platforms. 
It is possible that the extrinsic calibration of the cameras changed between datasets when the car doors were opened, or during driving if the mirrors experienced any flex. 
This is likely the reason the front-camera, which was more rigidly mounted, performed better on the real datasets. 
\begin{table}[h!]
  \centering
  \scriptsize
  \begin{tabularx}{\columnwidth}{l *{8}{Y}}
    \toprule
    & \multicolumn{8}{c}{APE Translation} \\
    \cmidrule{2-9}
    & \multicolumn{2}{c}{\textbf{MonoFront}} & \multicolumn{2}{c}{\textbf{MonoLeft}} & \multicolumn{2}{c}{\textbf{MonoRear}} & \multicolumn{2}{c}{\textbf{MonoRight}}  \\
Dataset &RMSE [m] & Drift [\%] & RMSE [m] & Drift [\%] & RMSE [m] & Drift [\%] & RMSE [m] & Drift [\%]  \\
 \midrule 
  sim1 (250m)  & 5.4                    & 2.2                    & \textcolor{green}{0.9} & \textcolor{green}{0.3} & 6.9    & 2.8  & 1.1                     & 0.4                    \\
  sim2 (468m)  & 1.5                    & 0.3                    & 1.4                    & 0.3                    & 2.7    & 0.6  & \textcolor{green}{0.9}  & \textcolor{green}{0.2} \\
  sim3 (749m)  & 6.1                    & 0.8                    & -                      & -                      & 4.7    & 0.6  & \textcolor{green}{3.3}  & \textcolor{green}{0.4} \\
  sim4 (804m)  & 7.1                    & 0.9                    & \textcolor{green}{5.5} & \textcolor{green}{0.7} & 13.3   & 1.7  & -                       & -                      \\
  \midrule
  out0 (498m)  & \textcolor{green}{0.5} & \textcolor{green}{0.1} & 1.3                    & 0.3                    & 1.9    & 0.4  & 6.1                     & 1.2                    \\
  out1 (445m)  & \textcolor{green}{1.6} & \textcolor{green}{0.4} & 1.8                    & 0.4                    & 4.7    & 1.0  & 4.0                     & 0.9                    \\
  out2 (521m)  & 1.9                    & 0.4                    & \textcolor{green}{1.4} & \textcolor{green}{0.3} & 195.6  & 37.5 & 3.3                     & 0.6                    \\
  out3 (807m)  & 7.2                    & 0.9                    & \textcolor{green}{3.8} & \textcolor{green}{0.5} & 10.7   & 1.3  & 7.2                     & 0.9                    \\
  out4 (537m)  & 1.5                    & 0.3                    & \textcolor{green}{0.5} & \textcolor{green}{0.1} & 5.1    & 0.9  & 1.3                     & 0.2                    \\
  out5 (514m)  & \textcolor{green}{1.3} & \textcolor{green}{0.2} & 2.0                    & 0.4                    & 1.7    & 0.3  & 29.4                    & 5.7                    \\
  out6 (448m)  & 3.1                    & 0.7                    & \textcolor{green}{2.6} & \textcolor{green}{0.6} & 4.5    & 1.0  & 3.0                     & 0.7                    \\
  out7 (406m)  & \textcolor{green}{1.2} & \textcolor{green}{0.3} & 2.1                    & 0.5                    & 24.0   & 5.9  & 3.9                     & 1.0                    \\
  out8 (48m)   & 1.5                    & 3.1                    & 1.0                    & 2.0                    & 8.9    & 2.5  & \textcolor{green}{0.3}  & \textcolor{green}{0.6} \\
  out9 (415m)  & 2.3                    & 0.5                    & 2.0                    & 0.5                    & 6.6    & 1.6  & \textcolor{green}{1.5}  & \textcolor{green}{0.4} \\
  out10 (486m) & 2.2                    & 0.4                    & \textcolor{green}{1.4} & \textcolor{green}{0.3} & 5.9    & 1.2  & 2.1                     & 0.4                    \\
  out11 (44m)  & 0.8                    & 1.8                    & \textcolor{green}{0.4} & \textcolor{green}{0.9} & 4.2    & 1.0  & 0.7                     & 1.5                    \\
  out12 (437m) & \textcolor{green}{1.2} & \textcolor{green}{0.3} & 1.4                    & 0.3                    & 162.4  & 37.2 & 2.5                     & 0.6                    \\
  out13 (341m) & \textcolor{green}{0.9} & \textcolor{green}{0.3} & 1.2                    & 0.4                    & 34.4   & 10.1 & 1.3                     & 0.4                    \\
  out14 (517m) & \textcolor{green}{1.2} & \textcolor{green}{0.2} & 1.3                    & 0.2                    & 6.9    & 1.3  & 1.7                     & 0.3                    \\
  out15 (194m) & \textcolor{green}{0.4} & \textcolor{green}{0.2} & 0.5                    & 0.3                    & 2.1    & 1.1  & 1.1                     & 0.6                    \\
  \midrule
  in0 (421m)   & 3.2                    & 0.8                    & 3.1                    & 0.7                    & 12.5   & 3.0  & \textcolor{green}{3.0}  & \textcolor{green}{0.7} \\
  in1 (321m)   & 4.8                    & 1.5                    & \textcolor{green}{3.5} & \textcolor{green}{1.1} & 5.5    & 1.7  & 4.3                     & 1.4                    \\
  in2 (563m)   & 12.2                   & 2.2                    & 65.8                   & 11.7                   & 13.4   & 2.4  & \textcolor{green}{10.0} & \textcolor{green}{1.8} \\
  in3 (416m)   & \textcolor{green}{4.2} & \textcolor{green}{1.0} & 11.7                   & 2.8                    & 6.5    & 1.6  & 11.1                    & 2.7                    \\
  in4 (723m)   & 22.8                   & 3.1                    & 9.7                    & 1.3                    & 24.7   & 3.4  & \textcolor{green}{8.9}  & \textcolor{green}{1.2} \\
  in5 (647m)   & 18.8                   & 2.9                    & 14.0                   & 2.2                    & 32.3   & 5.0  & \textcolor{green}{11.4} & \textcolor{green}{1.8} \\

    \bottomrule
  \end{tabularx}%
  \caption{VIO accuracy for each of the four cameras. 
  Best results for each dataset are highlighted in green. 
  Dashes are used to indicate tracking failures (drift > 100\%). 
  }
  \label{tab:kimera_mono_ate_real}
\vspace{-3mm}
\end{table}{}

Table \ref{tab:kimera_multicam_ate_real} shows the performance of the multi-camera configurations of Kimera. 
In the 1-camera configuration, only the left camera was used. 
For the 2-camera configuration, both the left and right cameras were used. 
Then we added the front camera, and finally the rear camera. 
The results highlighted in green are the best results, and those highlighted in blue are the second-best. 
Our expectation was that estimation error would decrease with each added camera, however this was not universally true. 
In the case of the simulated datasets, for 3 of the 4 datasets the best performance was found in the 4-camera configuration, while one dataset did best in the 2-camera configuration. 
However even in this case the 4-camera results were not far off. 
For the real datasets, the 1-camera configuration had the highest number of best-performance scores, though the other configurations also performed well in most cases. 
The degradation of performance with added sensors was likely due to the added error in extrinsic calibration with each camera, which could be partially alleviated through the use of extrinsic auto-calibration. 
In particular the 4-camera results for the real datasets were generally the poorest, which makes sense given the performance of the rear-camera in the monocular-VIO ablation study (Table \ref{tab:kimera_mono_ate_real}) were consistently subpar. 
\begin{table}[h!]
  \centering
  \scriptsize
  \begin{tabularx}{\columnwidth}{l *{8}{Y} | *2{Y}}
    \toprule
    & \multicolumn{10}{c}{APE Translation} \\
    \cmidrule{2-11}
    & \multicolumn{2}{c}{\textbf{1cam}}  & \multicolumn{2}{c}{\textbf{2cam}} & \multicolumn{2}{c}{\textbf{3cam}} & \multicolumn{2}{c}{\textbf{4cam}}  & \multicolumn{2}{c}{\textbf{1camWheel}} \\
Dataset & RMSE [m] & Drift [\%] & RMSE [m] & Drift [\%] & RMSE [m] & Drift [\%] & RMSE [m] & Drift [\%] & RMSE [m] & Drift [\%] \\
 \midrule 
  sim1   & 0.6                     & 0.2                         & 0.7                     & 0.3                         & \textcolor{blue}{0.6}   & \textcolor{blue}{0.2}       & \textcolor{green}{0.4} & \textcolor{green}{0.2} &               &               \\
  sim2   & 1.8                     & 0.4                         & \textcolor{blue}{1.4}   & \textcolor{blue}{0.3}       & 1.6                     & 0.3                         & \textcolor{green}{1.2} & \textcolor{green}{0.3} &               &               \\
  sim3   & 2.8                     & 0.4                         & 1.8                     & 0.2                         & \textcolor{blue}{1.5}   & \textcolor{blue}{0.2}       & \textcolor{green}{1.2} & \textcolor{green}{0.2} &               &               \\
  sim4   & 3.2                     & 0.4                         & \textcolor{green}{1.8}  & \textcolor{green}{0.2}      & 2.5                     & 0.3                         & \textcolor{blue}{1.9}  & \textcolor{blue}{0.2}  &               &               \\
  \midrule
  out0   & 6.1                     & 1.2                         & 8.4                     & 1.7                         & \textcolor{green}{2.4}  & \textcolor{green}{0.5}      & \textcolor{blue}{3.6}  & \textcolor{blue}{0.7}  & \textbf{1.8}  & \textbf{0.4}  \\
  out1   & 3.5                     & 0.8                         & \textcolor{green}{2.3}  & \textcolor{green}{0.5}      & \textcolor{blue}{3.4}   & \textcolor{blue}{0.8}       & 4.9                    & 1.1                    & \textbf{0.7}  & \textbf{0.2}  \\
  out2   & 3.3                     & 0.6                         & \textcolor{green}{2.0}  & \textcolor{green}{0.4}      & \textcolor{blue}{2.2}   & \textcolor{blue}{0.4}       & 2.6                    & 0.5                    & \textbf{0.7}  & \textbf{0.1}  \\
  out3   & 6.7                     & 0.8                         & \textcolor{blue}{2.9}   & \textcolor{blue}{0.4}       & \textcolor{green}{1.8}  & \textcolor{green}{0.2}      & 9.7                    & 1.2                    & \textbf{1.1}  & \textbf{0.1}  \\
  out4   & \textcolor{green}{1.3}  & \textcolor{green}{0.2}      & 4.6                     & 0.9                         & 5.8                     & 1.1                         & \textcolor{blue}{3.3}  & \textcolor{blue}{0.6}  & 1.5           & 0.3           \\
  out5   & \textcolor{green}{26.7} & \textcolor{green}{5.2}      & \textcolor{blue}{27.8}  & \textcolor{blue}{5.4}       & 31.9                    & 6.2                         & -                      & -                      & \textbf{1.7}  & \textbf{0.3}  \\
  out6   & \textcolor{green}{2.6}  & \textcolor{green}{0.6}      & \textcolor{blue}{4.0}   & \textcolor{blue}{0.9}       & 6.4                     & 1.4                         & 6.9                    & 1.5                    & \textbf{1.0}  & \textbf{0.2}  \\
  out7   & \textcolor{green}{3.6}  & \textcolor{green}{0.9}      & \textcolor{blue}{4.9}   & \textcolor{blue}{1.2}       & 8.8                     & 2.2                         & 6.0                    & 1.5                    & \textbf{1.2}  & \textbf{0.3}  \\
  out8   & \textcolor{green}{0.3}  & \textcolor{green}{0.6}      & 1.5                     & 3.1                         & \textcolor{blue}{0.4}   & \textcolor{blue}{0.9}       & 2.9                    & 6.1                    & \textbf{0.3}  & \textbf{0.6}  \\
  out9   & 2.0                     & 0.5                         & \textcolor{blue}{1.3}   & \textcolor{blue}{2.3}       & \textcolor{green}{0.7}  & \textcolor{green}{1.3}      & 3.9                    & 6.9                    & 1.2           & 0.3           \\
  out10  & 2.0                     & 0.4                         & \textcolor{blue}{1.8}   & \textcolor{blue}{0.4}       & \textcolor{green}{1.5}  & \textcolor{green}{0.3}      & 3.8                    & 0.8                    & \textbf{1.1}  & \textbf{0.2}  \\
  out11  & 0.7                     & 1.5                         & \textcolor{blue}{0.6}   & \textcolor{blue}{1.5}       & \textcolor{green}{0.2}  & \textcolor{green}{0.6}      & 1.0                    & 2.2                    & 0.3           & 0.7           \\
  out12  & \textcolor{green}{2.4}  & \textcolor{green}{0.6}      & \textcolor{blue}{4.3}   & \textcolor{blue}{1.0}       & 4.8                     & 1.1                         & 86.7                   & 19.8                   & \textbf{1.1}  & \textbf{0.2}  \\
  out13  & 1.5                     & 0.4                         & \textcolor{blue}{0.8}   & \textcolor{blue}{0.2}       & \textcolor{green}{0.5}  & \textcolor{green}{0.2}      & 1.1                    & 0.3                    & 0.9           & 0.3           \\
  out14  & \textcolor{blue}{1.4}   & \textcolor{blue}{0.3}       & \textcolor{green}{1.3}  & \textcolor{green}{0.3}      & 1.9                     & 0.4                         & 2.0                    & 0.4                    & \textbf{0.9}  & \textbf{0.2}  \\
  out15  & 1.1                     & 0.6                         & \textcolor{green}{0.7}  & \textcolor{green}{0.4}      & \textcolor{blue}{0.8}   & \textcolor{blue}{0.4}       & 1.0                    & 0.5                    & \textbf{0.2}  & \textbf{0.1}  \\
  \midrule
  in0    & \textcolor{green}{3.1}  & \textcolor{green}{0.7}      & \textcolor{blue}{28.4}  & \textcolor{blue}{6.7}       & 340.0                   & 80.7                        & 146.2                  & 34.7                   &               &              \\
  in1    & \textcolor{green}{4.4}  & \textcolor{green}{1.4}      & -                       & -                           & \textcolor{blue}{68.3}  & \textcolor{blue}{21.3}      & 103.4                  & 32.2                   &               &              \\
  in2    & \textcolor{green}{12.4} & \textcolor{green}{2.2}      & 56.9                    & 10.1                        & \textcolor{blue}{53.9}  & \textcolor{blue}{9.6}       & 68.5                   & 12.2                   &               &              \\
  in3    & \textcolor{green}{16.4} & \textcolor{green}{3.9}      & -                       & -                           & \textcolor{blue}{325.9} & \textcolor{blue}{78.2}      & 405.3                  & 97.2                   &               &              \\
  in4    & \textcolor{green}{16.4} & \textcolor{green}{2.3}      & \textcolor{blue}{217.7} & \textcolor{blue}{30.1}      & 691.7                   & 95.6                        & 645.6                  & 89.2                   &               &              \\
  in5    & \textcolor{green}{12.4} & \textcolor{green}{1.9}      & 322.3                   & 49.7                        & \textcolor{blue}{93.8}  & \textcolor{blue}{14.5}      & 115.4                  & 17.8                   &               &              \\
\bottomrule
\end{tabularx}%
\caption{Multi-camera VIO accuracy. 
Dataset length is omitted for brevity, see Tables \ref{tab:competitor_no_lc} or \ref{tab:kimera_mono_ate_real} for length.  
Best results are in green, second best in blue. 
Dashes are used to indicate tracking failures (drift > 100\%). 
The last column uses external odometry, results are boldfaced in cases where this is the best result. 
Wheel odometry was not present in simulated datasets or indoor datasets.}
\label{tab:kimera_multicam_ate_real}
\vspace{-3mm}
\end{table}{}

The final columns are walled off from the rest of the results as these were taken from the monocular VIO with the proprietary external odometry. 
In this case, wheel odometry was fed into Kimera-VIO and included in the backend as an odometry factor. 
This greatly improved estimation error, and the bolded results are the true best results of the table for the associated dataset. 
However these were separated from the rest of the table because comparing against pure VIO systems directly would have been inappropriate. 
Additionally, this data was not available in the simulated datasets or in indoor datasets. 
For the indoor datasets, wheel odometry was used in place of ground truth as the indoor scenarios were GPS-denied environments, so ground-truth was unavailable. 

Table \ref{tab:competitor_no_lc} shows the estimation performance of Kimera-VIO with one camera compared against two state of the art monocular VIO systems: Vins-Fusion \cite{Qin19arxiv-VINS-Fusion-odometry}, and Open-Vins \cite{Geneva20icra-openVINS}. 
ORB-SLAM3 is a SLAM-only pipeline so to turn off loop-closures and use it as a VIO-only pipeline would have been an unfair evaluation, so it was not included in this comparison. 
Results are shown for VIO systems without loop-closure. 
As these competitors do not support multiple surround-view cameras, 
the comparison is done in monocular mode for fairness. 
For each of the competitors, parameters were tuned for the best performance using a parameter regression script. 
For both Kimera and all competitor pipelines, the left camera was used for all datasets as it performed the most consistently in camera ablation studies on the competitor pipelines. 
Kimera-VIO out-performed both Vins-Fusion and Open-Vins in all but 4 datasets. 
Kimera-VIO struggled most with indoor scenes, though was within 0.3\% error from Vins-Fusion in all three of the indoor datasets in which it underperformed. 
Parameters for competitor pipelines are provided at \href{https://github.com/MIT-SPARK/ford-paper-params}{github.com/MIT-SPARK/ford-paper-params}. 

\begin{table}[h!]
  \vspace{-3mm}
  \centering
  \scriptsize
  \begin{tabularx}{\columnwidth}{l *{6}{Y}}
    \toprule
    & \multicolumn{6}{c}{\specialcell[b]{VIO Error (No Loop Closures) \\ APE Translation}} \\
    \cmidrule{2-7}
    & \multicolumn{2}{c}{\textbf{Kimera-1cam}} & \multicolumn{2}{c}{\textbf{Vins-Fusion}} & \multicolumn{2}{c}{\textbf{Open-Vins}} \\
Dataset &RMSE [m] & Drift [\%] & RMSE [m] & Drift [\%] & RMSE [m] & Drift [\%] \\
 \midrule 
 sim1 (250m)  & \textcolor{green}{0.6} & \textcolor{green}{0.2}        & -                       & -                            & 15.0                   & 6.0      \\
 sim2 (468m)  & \textcolor{green}{1.8} & \textcolor{green}{0.4}        & -                       & -                            & -                      & -        \\
 sim3 (749m)  & \textcolor{green}{2.8} & \textcolor{green}{0.4}        & -                       & -                            & 5.0                    & 0.7      \\
 sim4 (810m)  & \textcolor{green}{3.2} & \textcolor{green}{0.4}        & -                       & -                            & 17.0                   & 2.1      \\
 \midrule
 out0 (498m)  & \textcolor{green}{6.1}  & \textcolor{green}{1.2}       & -                       & -                            & 112.2                  & 22.6     \\
 out1 (445m)  & \textcolor{green}{3.5}  & \textcolor{green}{0.8}       & 202.9                   & 55.9                         & 419.7                  & 94.4     \\
 out2 (521m)  & \textcolor{green}{3.3}  & \textcolor{green}{0.6}       & -                       & -                            & 42.1                   & 8.1      \\
 out3 (807m)  & \textcolor{green}{6.7}  & \textcolor{green}{0.8}       & -                       & -                            & -                      & -        \\
 out4 (537m)  & \textcolor{green}{1.3}  & \textcolor{green}{0.2}       & -                       & -                            & -                      & -        \\
 out5 (514m)  & 26.7                    & 5.2                          & \textcolor{green}{12.1} & \textcolor{green}{3.2}       & 96.5                   & 18.8     \\
 out6 (448m)  & \textcolor{green}{2.6}  & \textcolor{green}{0.6}       & 3.3                     & 0.9                          & 17.0                   & 3.8      \\
 out7 (406m)  & \textcolor{green}{3.6}  & \textcolor{green}{0.9}       & 3.6                     & 1.3                          & 22.2                   & 5.5      \\
 out8 (48m)   & \textcolor{green}{0.3}  & \textcolor{green}{0.6}       & 10.2                    & 4.0                          & 18.1                   & 5.1      \\
 out9 (415m)  & \textcolor{green}{2.0}  & \textcolor{green}{0.5}       & 5.3                     & 1.8                          & 13.3                   & 3.2      \\
 out10 (486m) & \textcolor{green}{2.0}  & \textcolor{green}{0.4}       & 10.7                    & 2.9                          & 35.0                   & 7.2      \\
 out11 (44m)  & \textcolor{green}{0.7}  & \textcolor{green}{1.5}       & 4.7                     & 1.8                          & 20.3                   & 4.9      \\
 out12 (437m) & \textcolor{green}{2.4}  & \textcolor{green}{0.6}       & 8.0                     & 2.2                          & 45.1                   & 10.3     \\
 out13 (341m) & \textcolor{green}{1.5}  & \textcolor{green}{0.4}       & 2.7                     & 1.2                          & 15.1                   & 4.4      \\
 out14 (517m) & \textcolor{green}{1.4}  & \textcolor{green}{0.3}       & 4.7                     & 1.3                          & 33.5                   & 6.5      \\
 out15 (194m) & \textcolor{green}{1.1}  & \textcolor{green}{0.6}       & 1.9                     & 1.5                          & 7.2                    & 3.7      \\
 \midrule
 in0 (421m)   & \textcolor{green}{3.1}  & \textcolor{green}{0.7}       & 10.2                    & 3.5                          & 81.6                   & 19.3     \\
 in1 (321m)   & 4.4                     & 1.4                          & \textcolor{green}{2.6}  & \textcolor{green}{1.1}       & 79.8                   & 24.8     \\
 in2 (563m)   & 12.4                    & 2.2                          & \textcolor{green}{10.7} & \textcolor{green}{2.4}       & 97.2                   & 17.2     \\
 in3 (417m)   & \textcolor{green}{16.4} & \textcolor{green}{3.9}       & -                       & -                            & 63.1                   & 15.1     \\
 in4 (723m)   & 16.4                    & 2.3                          & \textcolor{green}{15.6} & \textcolor{green}{2.3}       & 191.1                  & 26.4     \\
 in5 (647m)   & \textcolor{green}{12.4} & \textcolor{green}{1.9}       & 13.2                    & 2.1                          & 420.5                  & 64.9     \\
 \bottomrule
  \end{tabularx}%
  \caption{VIO accuracy (no loop closures) of Kimera, Vins-Fusion, and Open-Vins.
  Best results for each dataset are highlighted in green. 
  Dashes are used to indicate tracking failures (drift > 100\%). 
  }
  \label{tab:competitor_no_lc}
  \vspace{-6mm}
\end{table}{}

\subsection{Loop-Closure Detection}
\label{subsec:lc-detection}

The proposed system provides several schemes for closing the SLAM loop by performing loop-closures on the monocular image data. 
As the cameras had very little image overlap, it was not feasible to use the standard stereo-matching methods to generate depth data for calculating relative poses between loop closure match candidates as in~\cite{Rosinol21ijrr-Kimera,Rosinol20icra-Kimera}. 
Table \ref{tab:lcd_ablation} shows the results of an ablation study on these various methods for performing loop closures. 
The impact of each method on the estimation error and drift of the SLAM system are shown, with best results highlighted in green. 
The first pair of columns are for the VIO system without loop closures (``VIO''). 
The remaining columns represent the three loop-closure system described in section~\ref{sec:systemarchitecture}. 
Only datasets with loops are included in the analysis. 
The proposed scale-less factor was far superior to the other methods and to simple VIO except in a few select circumstances, and even in those cases RPGO was able to reject bad loop closure candidates and prevent the estimate from being worse than the simple VIO estimate. 
In some cases this method was able to reduce drift by a factor of four, and in most datasets we were able to obtain drift less than 1\% of the trajectory length. 
Table~\ref{tab:competitor_lc} reports the VIO performance of Kimera (monocular) with loop closures against Vins-Fusion with Loop-Fusion \cite{Qin19arxiv-VINS-Fusion-odometry} and ORB-SLAM3 \cite{Campos21-TRO}. 
Kimera beats the other competitors in all of the datasets in this case. 


\begin{figure}[h]
\vspace{-3mm}
\centering
\begin{subfigure}{\columnwidth}
    \centering
    \includegraphics[width=\columnwidth]{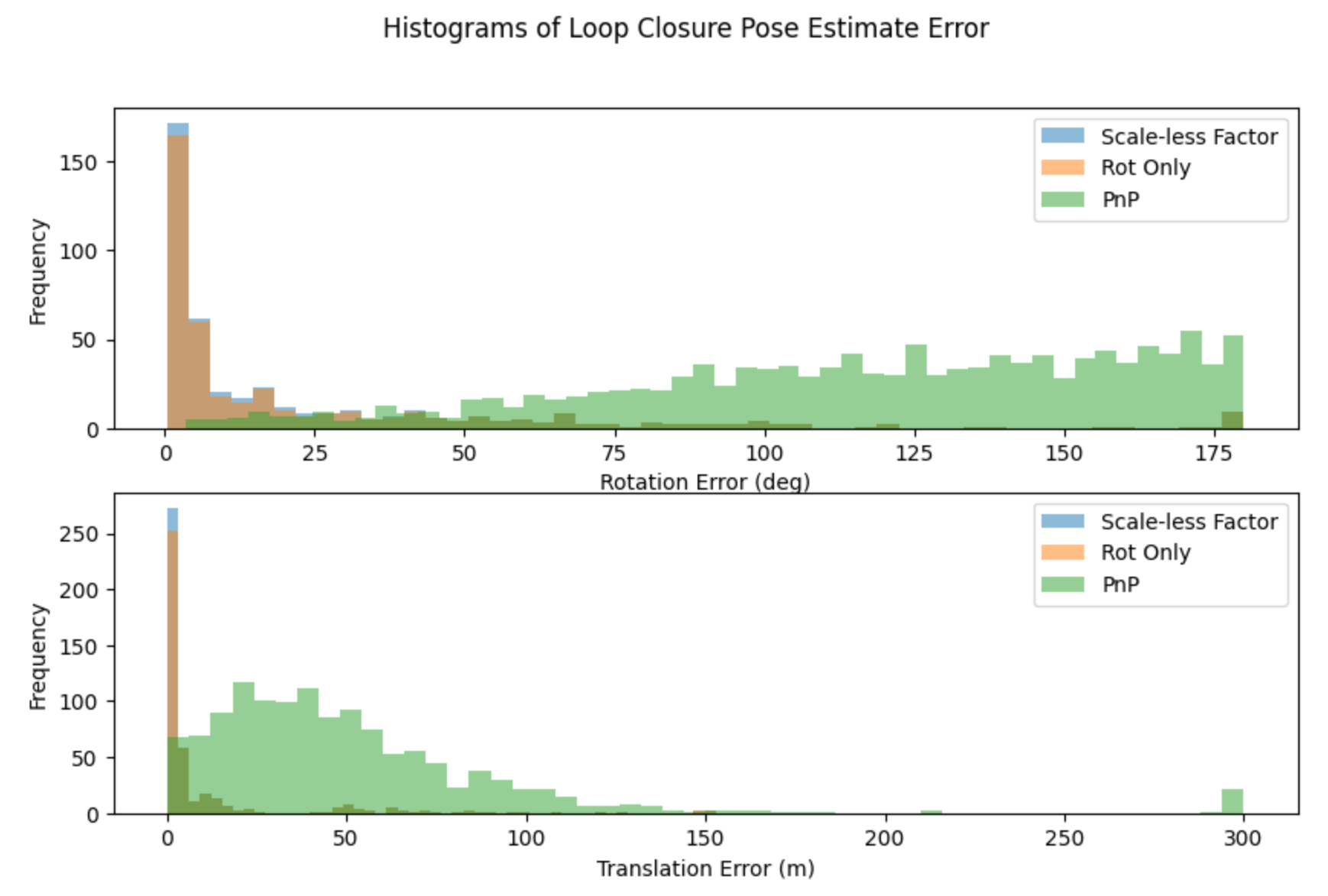}
\end{subfigure}%
\caption{Histograms of the proposed loop-closure detection methods. 
Each bin is error on rotation and translation, and contains the sum total of all loop-closure candidates across all datasets that scored within that bin. 
}
\label{fig:lcd-histograms}
\vspace{-3mm}
\end{figure}

\begin{table}[h!]
  \centering
  \scriptsize
  \begin{tabularx}{\columnwidth}{l *{8}{Y}}
    \toprule
    & \multicolumn{8}{c}{\specialcell[b]{Loop Closure Detection Ablation Study \\ APE Translation}} \\
    \cmidrule{2-9}
    & \multicolumn{2}{c}{\textbf{VIO}} & \multicolumn{2}{c}{\textbf{Scale-less Factor}} & \multicolumn{2}{c}{\textbf{Rot Only}} & \multicolumn{2}{c}{\textbf{PnP}}  \\ 
Dataset &RMSE [m] & Drift [\%] & RMSE [m] & Drift [\%] & RMSE [m] & Drift [\%] & RMSE [m] & Drift [\%] \\ 
 \midrule 
  sim2 (468m)  & 1.8  & 0.4  & \textcolor{green}{0.5}  & \textcolor{green}{0.1} & 1.1                     & 0.2                     & 0.8                     & 0.2                     \\ 
  sim3 (749m)  & 2.8  & 0.4  & \textcolor{green}{2.5}  & \textcolor{green}{0.3} & 4.0                     & 0.5                     & 3.3                     & 0.4                     \\ 
  sim4 (804m)  & 3.2  & 0.4  & 3.2                     & 0.4                    & 3.2                     & 0.4                     & 3.2                     & 0.4                     \\ 
  \midrule
  out6 (448m)  & 2.6  & 0.6  & 2.6                     & 0.6                    & 2.6                     & 0.6                     & 2.6                     & 0.6                     \\ 
  out7 (290m)  & 9.7  & 2.4  & \textcolor{green}{3.9}  & \textcolor{green}{1.0} & 7.7                     & 2.7                     & 13.0                    & 3.2                     \\ 
  out8 (48m)   & 0.4  & 0.9  & 0.4                     & 0.9                    & \textcolor{green}{0.2}  & \textcolor{green}{0.5}  & 0.3                     & 0.7                     \\ 
  out9 (415m)  & 1.6  & 0.4  & \textcolor{green}{1.3}  & \textcolor{green}{0.3} & 2.0                     & 0.5                     & 4.0                     & 1.0                     \\ 
  out10 (486m) & 2.4  & 0.5  & \textcolor{green}{1.6}  & \textcolor{green}{0.3} & 2.6                     & 0.5                     & 2.2                     & 0.4                     \\ 
  out11 (44m)  & 0.7  & 1.5  & \textcolor{green}{0.6}  & \textcolor{green}{1.5} & 0.7                     & 1.5                     & 0.7                     & 1.5                     \\ 
  out12 (437m) & 2.4  & 0.6  & \textcolor{green}{1.9}  & \textcolor{green}{0.4} & 2.6                     & 0.6                     & 6.4                     & 1.5                     \\ 
  out13 (341m) & 1.5  & 0.4  & 1.2                     & 0.3                    & 1.6                     & 0.5                     & \textcolor{green}{1.1}  & \textcolor{green}{0.3}  \\ 
  out14 (517m) & 1.4  & 0.3  & \textcolor{green}{1.1}  & \textcolor{green}{0.2} & 1.2                     & 0.2                     & 4.0                     & 0.8                     \\ 
  \midrule
  in0 (421m)   & 3.1  & 0.7  & \textcolor{green}{2.7}  & \textcolor{green}{0.7} & 3.1                     & 0.7                     & 5.5                     & 1.3                     \\ 
  in1 (321m)   & 4.4  & 1.4  & \textcolor{green}{2.0}  & \textcolor{green}{0.6} & 5.4                     & 1.7                     & 3.1                     & 1.0                     \\ 
  in2 (563m)   & 12.4 & 2.2  & 12.4                    & 2.2                    & \textcolor{green}{12.4} & \textcolor{green}{2.2}  & 12.7                    & 2.3                     \\ 
  in3 (417m)   & 16.5 & 4.0  & \textcolor{green}{14.1} & \textcolor{green}{3.4} & 16.4                    & 3.9                     & 16.4                    & 3.9                     \\ 
  in4 (723m)   & 16.4 & 2.3  & \textcolor{green}{12.8} & \textcolor{green}{1.8} & 16.3                    & 2.3                     & 16.4                    & 2.3                     \\ 
  in5 (647m)   & 12.3 & 1.9  & 12.3                    & 1.9                    & 12.4                    & 1.9                     & \textcolor{green}{12.4} & \textcolor{green}{1.9}  \\ 
 \bottomrule
  \end{tabularx}%
  \caption{Pose estimation accuracy (including loop closures) restricted to datasets that contain loops. First column is VIO only (no loop closures), and all configurations use only 1 camera.}
  \label{tab:lcd_ablation}
  \vspace{-2mm}
\end{table}{}


\begin{figure*}[h]
\vspace{-4mm}
\centering
\begin{subfigure}{\textwidth}
    \centering
    \includegraphics[width=\textwidth]{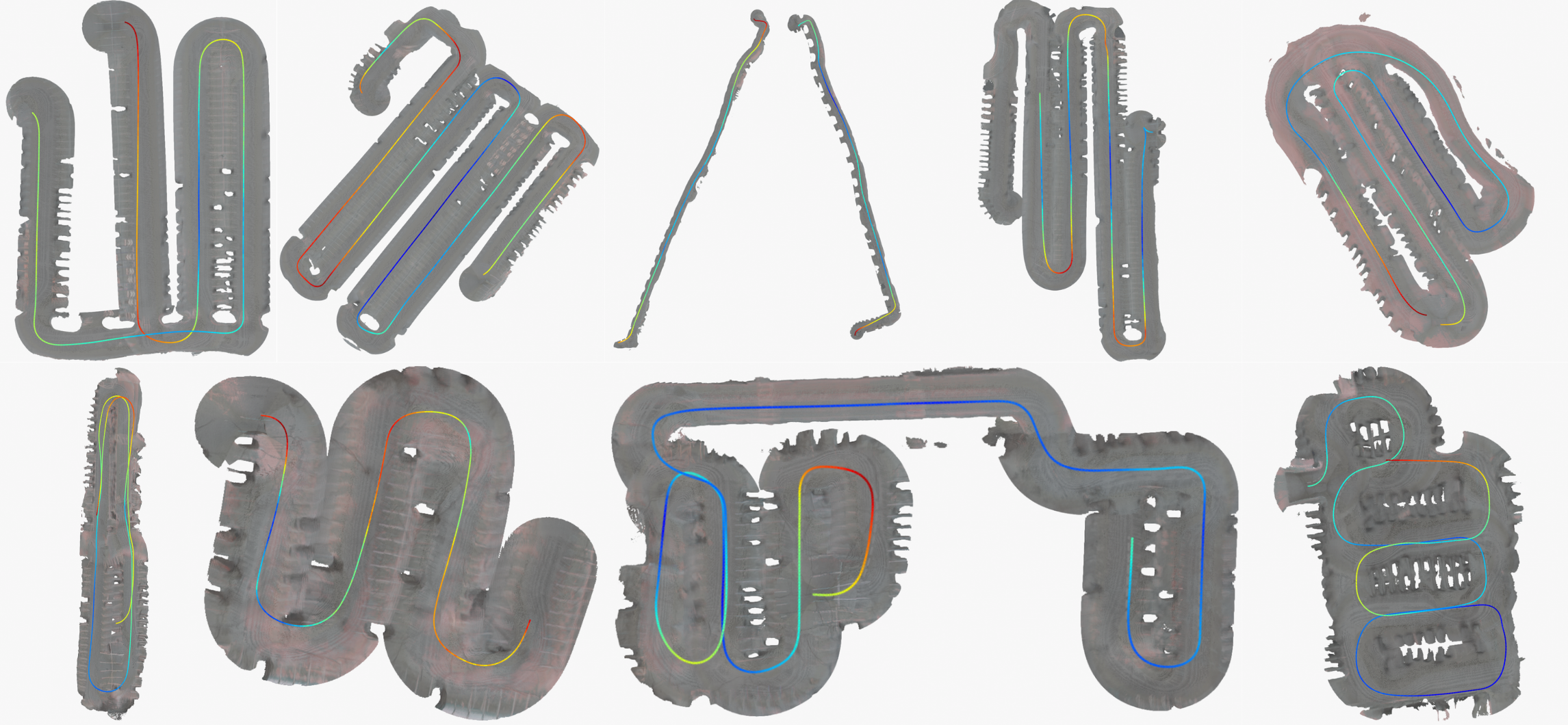}
\end{subfigure}%
\caption{3D reconstructions produced by the proposed free-space mapping approach on several Ford datasets.
All four cameras were used for reconstruction, and Kimera's visual-inertial odometry was performed with all four cameras and external odometry.
A colormap of the estimated trajectory is plotted over each reconstruction, with cooler colors representing lower ATE RMSE. 
}
\label{fig:ford-reconstructions}
\vspace{-5mm}
\end{figure*}

Figure \ref{fig:lcd-histograms} compares histograms of the rotation and translation errors for loop closure candidates generated by all three methods. 
It is clear that the scale-less approach has the lowest errors consistently, and that the PnP method failed to get accurate relative poses. 
In the PnP method, RANSAC found on average of 5 inliers across all candidates across all datasets. 
This is likely due to a mismatch between the detected ORB keypoints and the backend-tracked 3D landmarks. 
On the other hand, both the scale-less and the rotation-only approaches had upwards of 10 inliers in average. 


\begin{table}[h!]
  \centering
  \scriptsize
  \begin{tabularx}{\columnwidth}{l *{6}{Y}}
    \toprule
    & \multicolumn{6}{c}{\specialcell[b]{VI-SLAM Comparison With Loop Closures \\ APE Translation}} \\
    \cmidrule{2-7}
    & \multicolumn{2}{c}{\textbf{Kimera-1cam}} & \multicolumn{2}{c}{\textbf{Vins-Fusion}} & \multicolumn{2}{c}{\textbf{ORB-SLAM3}}  \\
Dataset &RMSE [m] & Drift [\%] & RMSE [m] & Drift [\%] & RMSE [m] & Drift [\%] \\
 \midrule 
 sim2 (468m)  & \textcolor{green}{0.5}  & \textcolor{green}{0.1}       & -       & -             & 22.6 & 4.8    \\
 sim3 (749m)  & \textcolor{green}{2.5}  & \textcolor{green}{0.3}       & -       & -             & 61.2 & 8.2    \\
 sim4 (810m)  & \textcolor{green}{3.2}  & \textcolor{green}{0.4}       & -       & -             & 15.3 & 1.9    \\
 \midrule
 out6 (448m)  & \textcolor{green}{2.6}  & \textcolor{green}{0.6}       & 3.4     & 0.8           & 37.8 & 8.4    \\
 out7 (406m)  & \textcolor{green}{3.9}  & \textcolor{green}{1.0}       & 4.1     & 1.0           & 8.9  & 2.2    \\
 out8 (48m)   & \textcolor{green}{0.4}  & \textcolor{green}{0.9}       & 12.2    & 3.4           & 12.7 & 3.6    \\
 out9 (415m)  & \textcolor{green}{1.3}  & \textcolor{green}{0.3}       & 5.2     & 1.3           & 6.4  & 1.5    \\
 out10 (486m) & \textcolor{green}{1.6}  & \textcolor{green}{0.3}       & 14.3    & 2.9           & 10.5 & 2.2    \\
 out11 (44m)  & \textcolor{green}{0.6}  & \textcolor{green}{1.5}       & 7.9     & 1.9           & 9.0  & 2.2    \\
 out12 (437m) & \textcolor{green}{1.9}  & \textcolor{green}{0.4}       & 7.5     & 1.7           & 1.9  & 1.8    \\
 out13 (341m) & \textcolor{green}{1.2}  & \textcolor{green}{0.3}       & 3.3     & 1.0           & 11.9 & 3.5    \\
 out14 (517m) & \textcolor{green}{1.1}  & \textcolor{green}{0.2}       & 5.3     & 1.0           & 20.9 & 4.0    \\
 \midrule
 in0 (421m)   & \textcolor{green}{2.7}  & \textcolor{green}{0.7}       & 10.4    & 2.5           & 3.6  & 0.8    \\
 in1 (321m)   & \textcolor{green}{2.0}  & \textcolor{green}{0.6}       & 2.7     & 0.8           & 7.2  & 2.2    \\
 in2 (563m)   & \textcolor{green}{12.4} & \textcolor{green}{2.2}       & 13.2    & 2.3           & 43.4 & 15.4   \\
 in3 (417m)   & \textcolor{green}{14.1} & \textcolor{green}{3.4}       & -       & -             & 48.1 & 11.5   \\
 in4 (723m)   & \textcolor{green}{12.8} & \textcolor{green}{1.8}       & 16.3    & 2.3           & 50.6 & 7.0    \\
 in5 (647m)   & \textcolor{green}{12.3} & \textcolor{green}{1.9}       & 15.3    & 2.4           & 33.0 & 5.1    \\
 \bottomrule
  \end{tabularx}%
  \caption{Pose estimation accuracy (including loop closures) for Kimera, Vins-Fusion, and ORB-SLAM3.
  Best results for each dataset are highlighted in green. 
  Dashes are used to indicate tracking failures (drift > 100\%). 
  } 
  \label{tab:competitor_lc}
  \vspace{-1mm}
\end{table}{}

\subsection{Ground Plane Reconstruction}
\label{subsec:free-space-recon}

Table~\ref{tab:tesse_reconstruction_eval} shows geometric reconstruction accuracy (as defined in~\cite{Rosinol21ijrr-Kimera}) for the simulated datasets, where ground-truth ground-plane maps are available. 
The third column of Table \ref{tab:tesse_reconstruction_eval} shows the reconstruction accuracy when using the RPGO trajectory: more precisely, in this case we use the pose graph and mesh optimization approach in~\cite{Rosinol21ijrr-Kimera} to jointly optimize the mesh and trajectory. 
Some additional trajectory error comes from using Kimera's poses in all of the simulated datasets, which is expected. 
However, we observe that the results in the last column remain close to the ones obtained with ground-truth poses.




\begin{table}[h!]
  \centering
  \scriptsize
  \begin{tabularx}{\columnwidth}{l *{3}{Y}}
    \toprule
    & \multicolumn{3}{c}{\specialcell[b]{Kimera-Semantics Geometric Reconstruction Accuracy \\ ATE RMSE [m]}} \\
    \cmidrule{2-4}
    Dataset &\multicolumn{1}{c}{\specialcell[b]{Homography \\ GT Poses}} & \multicolumn{1}{c}{\specialcell[b]{Homography \\ Kimera-VIO Poses}} & \multicolumn{1}{c}{\specialcell[b]{Homography \\ Kimera-RPGO Poses}}  \\
    \midrule 
    sim1 (250m) & 0.22 & 0.26 & 0.22 \\
    sim2 (468m) & 0.37 & 0.40 & 0.37 \\
    sim3 (749m) & 0.23 & 0.32 & 0.29 \\
    sim4 (810m) & 0.29 & 0.35 & 0.34 \\
    \bottomrule
  \end{tabularx}%
  \caption{Geometric reconstruction accuracy for the modified Kimera-Semantics using three different configurations.}
  \label{tab:tesse_reconstruction_eval}
\end{table}{}

Figure \ref{fig:ford-reconstructions} shows several free-space reconstructions of the Ford datasets using 
the proposed homography-based method. 
The maps were generated using Kimera-VIO's pose with 4-cameras and external odometry.



\section{Conclusions}
\label{sec:conclusions}

We proposed modifications to Kimera to support multi-camera input data, and to 
incorporate external odometry inputs.
To complete the SLAM system, we modified the loop-closure module to utilize monocular inputs. 
Kimera-Semantics was modified to perform free-space mapping for autonomous valet parking applications, 
which works with monocular cameras and does not require learning for depth-estimation.
We tested the system on simulated car data and real-world datasets collected with a test car at the Ford Motor Company. 
The proposed system exhibits small trajectory and mapping errors and consistently outperforms state-of-the-art open-source VIO and VI-SLAM systems. 
Real-world results for the multi-camera system also suggest room for improvement, 
in particular with respect to extrinsic calibration. 
Automatic calibration in the backend factor graph could alleviate some of the issues. 
Additionally, Kimera-Semantics supports semantic-annotations for any number of object classes; 
therefore, a potential extension  could include mapping other semantic classes (\eg parking spots, pedestrians) 
relevant to  autonomous parking. 

\isExtended{}{\vspace{-2mm}}
\bibliographystyle{IEEEtran}
\bibliography{../references/refs,../references/myRefs}

\begin{thebibliography}{10}
\providecommand{\url}[1]{#1}
\csname url@samestyle\endcsname
\providecommand{\newblock}{\relax}
\providecommand{\bibinfo}[2]{#2}
\providecommand{\BIBentrySTDinterwordspacing}{\spaceskip=0pt\relax}
\providecommand{\BIBentryALTinterwordstretchfactor}{4}
\providecommand{\BIBentryALTinterwordspacing}{\spaceskip=\fontdimen2\font plus
\BIBentryALTinterwordstretchfactor\fontdimen3\font minus
  \fontdimen4\font\relax}
\providecommand{\BIBforeignlanguage}[2]{{%
\expandafter\ifx\csname l@#1\endcsname\relax
\typeout{** WARNING: IEEEtran.bst: No hyphenation pattern has been}%
\typeout{** loaded for the language `#1'. Using the pattern for}%
\typeout{** the default language instead.}%
\else
\language=\csname l@#1\endcsname
\fi
#2}}
\providecommand{\BIBdecl}{\relax}
\BIBdecl

\bibitem{Forster17tro}
C.~Forster, L.~Carlone, F.~Dellaert, and D.~Scaramuzza, ``On-manifold
  preintegration for real-time visual-inertial odometry,'' \emph{{IEEE} Trans.
  Robotics}, vol.~33, no.~1, pp. 1--21, 2017, arxiv preprint: 1512.02363,
  \linkToPdf{http://arxiv.org/abs/1512.02363}, technical report
  GT-IRIM-CP\&R-2015-001\award{, Transactions on Robotics Best paper award}.

\bibitem{Mourikis09tro-EdlSoundingRocket}
A.~I. Mourikis, N.~Trawny, S.~I. Roumeliotis, A.~E. Johnson, A.~Ansar, and
  L.~Matthies, ``Vision-aided inertial navigation for spacecraft entry,
  descent, and landing,'' \emph{IEEE Transactions on Robotics}, vol.~25, no.~2,
  pp. 264--280, 2009.

\bibitem{Cadena16tro-SLAMsurvey}
C.~Cadena, L.~Carlone, H.~Carrillo, Y.~Latif, D.~Scaramuzza, J.~Neira, I.~Reid,
  and J.~Leonard, ``Past, present, and future of simultaneous localization and
  mapping: Toward the robust-perception age,'' \emph{{IEEE} Trans. Robotics},
  vol.~32, no.~6, pp. 1309--1332, 2016, arxiv preprint: 1606.05830,
  \linkToPdf{https://arxiv.org/abs/1606.05830}.

\bibitem{Shao20acm-ATS}
X.~Shao, L.~Zhang, T.~Zhang, Y.~Shen, H.~Li, and Y.~Zhou, ``A tightly-coupled
  semantic slam system with visual, inertial and surround-view sensors for
  autonomous indoor parking,'' \emph{Proceedings of the 28th ACM International
  Conference on Multimedia}, 2020.

\bibitem{Shao22acm-SLAMFI}
X.~Shao, Y.~Shen, L.~Zhang, S.~Zhao, D.~Zhu, and Y.~Zhou, ``Slam for indoor
  parking: A comprehensive benchmark dataset and a tightly coupled semantic
  framework,'' \emph{ACM Transactions on Multimedia Computing, Communications
  and Applications}, vol.~19, pp. 1 -- 23, 2022.

\bibitem{Yu22ieee-HierarchicalMI}
J.~Yu, Z.-Z. Xiang, and J.~Su, ``Hierarchical multi-level information fusion
  for robust and consistent visual slam,'' \emph{IEEE Transactions on Vehicular
  Technology}, vol.~71, pp. 250--259, 2022.

\bibitem{Xiang21icra-HybridBirdsEyeValet}
Z.~Xiang, A.~Bao, and J.~Su, ``Hybrid bird’s-eye edge based semantic visual
  slam for automated valet parking,'' in \emph{2021 IEEE International
  Conference on Robotics and Automation (ICRA)}, 2021, pp. 11\,546--11\,552.

\bibitem{Khoche22icits-Semantic3G}
A.~Khoche, M.~K. Wozniak, D.~Duberg, and P.~Jensfelt, ``Semantic 3d grid maps
  for autonomous driving,'' \emph{2022 IEEE 25th International Conference on
  Intelligent Transportation Systems (ITSC)}, pp. 2681--2688, 2022.

\bibitem{Shi22-VIPSRP}
S.~Shi, J.~Cui, Z.~Jiang, Z.~Yan, G.~Xing, J.~Niu, and Z.~Ouyang, ``Vips:
  real-time perception fusion for infrastructure-assisted autonomous driving,''
  \emph{Proceedings of the 28th Annual International Conference on Mobile
  Computing And Networking}, 2022.

\bibitem{Rosinol21ijrr-Kimera}
A.~Rosinol, A.~Violette, M.~Abate, N.~Hughes, Y.~Chang, J.~Shi, A.~Gupta, and
  L.~Carlone, ``Kimera: from {SLAM} to spatial perception with {3D} dynamic
  scene graphs,'' \emph{Intl. J. of Robotics Research}, vol.~40, no. 12--14,
  pp. 1510--1546, 2021, arXiv preprint: 2101.06894,
  \linkToPdf{https://arxiv.org/pdf/2101.06894.pdf}.

\bibitem{Rosinol20icra-Kimera}
A.~Rosinol, M.~Abate, Y.~Chang, and L.~Carlone, ``Kimera: an open-source
  library for real-time metric-semantic localization and mapping,'' in
  \emph{IEEE Intl. Conf. on Robotics and Automation (ICRA)}, 2020, arXiv
  preprint: 1910.02490,
  \linkToVideo{https://www.youtube.com/watch?v=-5XxXRABXJs},
  \linkToCode{https://github.com/MIT-SPARK/Kimera},
  \linkToPdf{https://arxiv.org/pdf/1910.02490.pdf}.

\bibitem{Qin18tro-vinsmono}
T.~Qin, P.~Li, and S.~Shen, ``Vins-mono: A robust and versatile monocular
  visual-inertial state estimator,'' \emph{IEEE Transactions on Robotics},
  vol.~34, no.~4, pp. 1004--1020, 2018.

\bibitem{Mur-Artal17tro-ORBSLAM2}
R.~Mur-Artal and J.~D. Tard\'os, ``{ORB-SLAM2}: an open-source {SLAM} system
  for monocular, stereo and {RGB-D} cameras,'' \emph{{IEEE} Trans. Robotics},
  vol.~33, no.~5, pp. 1255--1262, 2017.

\bibitem{Qin19arxiv-VINS-Fusion-poseEstimation}
T.~Qin, S.~Cao, J.~Pan, and S.~Shen, ``A general optimization-based framework
  for global pose estimation with multiple sensors,'' \emph{arXiv preprint:
  1901.03642}, 2019.

\bibitem{Campos21-TRO}
C.~Campos, R.~Elvira, J.~J.~G. Rodr{\'\i}guez, J.~M. Montiel, and J.~D.
  Tard{\'o}s, ``{ORB-SLAM3}: An accurate open-source library for visual,
  visual--inertial, and multimap {SLAM},'' \emph{{IEEE} Trans. Robotics}, 2021.

\bibitem{Geneva20icra-openVINS}
P.~Geneva, K.~Eckenhoff, W.~Lee, Y.~Yang, and G.~Huang, ``{OpenVINS}: A
  research platform for visual-inertial estimation,'' in \emph{IEEE Intl. Conf.
  on Robotics and Automation (ICRA)}, 2020, pp. 4666--4672.

\bibitem{OpenCV}
G.~Bradski, ``{The OpenCV Library},'' \emph{Dr. Dobb's Journal of Software
  Tools}, 2000.

\bibitem{gtsam}
{F. Dellaert et al.}, ``{Georgia Tech Smoothing And Mapping (GTSAM)},''
  \url{https://gtsam.org/}, 2019.

\bibitem{Agarwal12-ceres}
S.~Agarwal, K.~Mierle \emph{et~al.}, ``Ceres solver,'' 2012.

\bibitem{Chang22ral-LAMP2}
Y.~Chang, K.~Ebadi, C.~Denniston, M.~F. Ginting, A.~Rosinol, A.~Reinke,
  M.~Palieri, J.~Shi, C.~A, B.~Morrell, A.~Agha-mohammadi, and L.~Carlone,
  ``{LAMP 2.0}: A robust multi-robot {SLAM} system for operation in challenging
  large-scale underground environments,'' \emph{{IEEE} Robotics and Automation
  Letters ({RA-L})}, vol.~7, no.~4, pp. 9175--9182, 2022,
  \linkToPdf{https://arxiv.org/pdf/2205.13135.pdf}.

\bibitem{Eckenhoff20arxiv-MIMICVINS}
\BIBentryALTinterwordspacing
K.~Eckenhoff, P.~Geneva, and G.~Huang, ``Mimc-vins: A versatile and resilient
  multi-imu multi-camera visual-inertial navigation system,'' 2020. [Online].
  Available: \url{https://arxiv.org/abs/2006.15699}
\BIBentrySTDinterwordspacing

\bibitem{Yang21arxiv-AsyncMultiViewSLAM}
\BIBentryALTinterwordspacing
A.~J. Yang, C.~Cui, I.~A. Bârsan, R.~Urtasun, and S.~Wang, ``Asynchronous
  multi-view slam,'' 2021. [Online]. Available:
  \url{https://arxiv.org/abs/2101.06562}
\BIBentrySTDinterwordspacing

\bibitem{Zhang21ral-BalancingTB}
L.~Zhang, D.~Wisth, M.~Camurri, and M.~F. Fallon, ``Balancing the budget:
  Feature selection and tracking for multi-camera visual-inertial odometry,''
  \emph{{IEEE} Robotics and Automation Letters}, vol.~7, pp. 1182--1189, 2021.

\bibitem{Zhang22irs_MMOSlam}
Z.~Zhang and K.~Zou, ``Mmo-slam: A versatile and accurate multi monocular slam
  system,'' \emph{Journal of Intelligent and Robotic Systems}, vol. 105, no.
  1573-0409, 2022.

\bibitem{He21arxiv-TowardEA}
Y.~He, H.~Yu, W.~Yang, and S.~A. Scherer, ``Toward efficient and robust
  multiple camera visual-inertial odometry,'' \emph{ArXiv}, vol.
  abs/2109.12030, 2021.

\bibitem{Wang20icra-ReliableFM}
Y.~Wang, K.~Huang, X.-Z. Peng, H.~Li, and L.~Kneip, ``Reliable frame-to-frame
  motion estimation for vehicle-mounted surround-view camera systems,''
  \emph{2020 IEEE International Conference on Robotics and Automation (ICRA)},
  pp. 1660--1666, 2020.

\bibitem{Tripathi20arxiv-TrainedParkingSurroundView}
\BIBentryALTinterwordspacing
N.~Tripathi and S.~Yogamani, ``Trained trajectory based automated parking
  system using visual slam on surround view cameras,'' 2020. [Online].
  Available: \url{https://arxiv.org/abs/2001.02161}
\BIBentrySTDinterwordspacing

\bibitem{Shao22CSVT-MOFISSLAMAM}
X.~Shao, L.~Zhang, T.~Zhang, Y.~Shen, and Y.~Zhou, ``Mofisslam: A multi-object
  semantic slam system with front-view, inertial, and surround-view sensors for
  indoor parking,'' \emph{IEEE Transactions on Circuits and Systems for Video
  Technology}, vol.~32, pp. 4788--4803, 2022.

\bibitem{Wu22ieee-VinsDynamicObjDet}
X.~Wu, F.~Huang, Y.~Wang, and H.~Jiang, ``A vins combined with dynamic object
  detection for autonomous driving vehicles,'' \emph{IEEE Access}, vol.~10, pp.
  91\,127--91\,136, 2022.

\bibitem{Wimbauer20arxiv-MonoRec}
\BIBentryALTinterwordspacing
F.~Wimbauer, N.~Yang, L.~von Stumberg, N.~Zeller, and D.~Cremers, ``Monorec:
  Semi-supervised dense reconstruction in dynamic environments from a single
  moving camera,'' 2020. [Online]. Available:
  \url{https://arxiv.org/abs/2011.11814}
\BIBentrySTDinterwordspacing

\bibitem{Forster15rss-imuPreintegration}
C.~Forster, L.~Carlone, F.~Dellaert, and D.~Scaramuzza, ``{IMU} preintegration
  on manifold for efficient visual-inertial maximum-a-posteriori estimation,''
  in \emph{Robotics: Science and Systems (RSS)}, 2015, accepted as oral
  presentation (acceptance rate $4\%$)
  \linkToPdf{http://www.roboticsproceedings.org/rss11/p06.html}
  \linkToVideo{https://www.youtube.com/watch?v=CsJkci5lfco} (supplemental
  material:
  \linkToPdf{https://www.dropbox.com/s/kzraqftn22bjb0u/2015c-RSS-VIN-supplementaryMaterial.pdf?dl=0})\award{,
  finalist for best paper award}.

\bibitem{Scaramuzza06iros-omniCamera}
D.~Scaramuzza, A.~Martinelli, and R.~Siegwart, ``A toolbox for easy calibrating
  omnidirectional cameras,'' in \emph{IEEE/RSJ Intl. Conf. on Intelligent
  Robots and Systems (IROS)}, 2006.

\bibitem{Galvez12tro-dbow}
D.~G\'alvez-L\'opez and J.~D. Tard\'os, ``Bags of binary words for fast place
  recognition in image sequences,'' \emph{IEEE Transactions on Robotics},
  vol.~28, no.~5, pp. 1188--1197, October 2012.

\bibitem{Yang20ral-GNC}
H.~Yang, P.~Antonante, V.~Tzoumas, and L.~Carlone, ``Graduated non-convexity
  for robust spatial perception: From non-minimal solvers to global outlier
  rejection,'' \emph{{IEEE} Robotics and Automation Letters ({RA-L})}, vol.~5,
  no.~2, pp. 1127--1134, 2020, arXiv preprint:1909.08605 (with supplemental
  material), \linkToPdf{https://arxiv.org/pdf/1909.08605.pdf}\award{, ICRA Best
  paper award in Robot Vision}.

\bibitem{Hartley00}
R.~Hartley and A.~Zisserman, \emph{Multiple View Geometry in Computer
  Vision}.\hskip 1em plus 0.5em minus 0.4em\relax Cambridge University Press,
  2000.

\bibitem{Qin19arxiv-VINS-Fusion-odometry}
T.~Qin, J.~Pan, S.~Cao, and S.~Shen, ``A general optimization-based framework
  for local odometry estimation with multiple sensors,'' \emph{arXiv preprint:
  1901.03638}, 2019.

\end{thebibliography}

\end{document}